%% file: main.tex
\documentclass[11pt,a4paper]{article}

\usepackage{times,latexsym}
\usepackage[T1]{fontenc}
\usepackage{url}
\usepackage{amsmath,amssymb}
\usepackage{booktabs}
\usepackage{array}
\usepackage{graphicx}
\graphicspath{{figures/}}
\usepackage{float}
\usepackage{microtype}
\usepackage[acceptedWithA]{tacl2021v1}

\hypersetup{
  pdftitle={What Fixed-Rollout pass@k Evaluations Can Identify},
  pdfauthor={Pranav Singh, Prashant Singh},
  pdfsubject={Statistical identification limits of fixed-rollout pass@k evaluation},
  pdfkeywords={pass@k, partial identification, moment problem, LLM evaluation,
    inference-time scaling}
}

\newtheorem{theorem}{Theorem}
\title{What Fixed-Rollout pass@\(k\) Evaluations Can Identify}
\author{Pranav Singh \\
  Department of Mathematics \\
  Indian Institute of Technology Ropar \\
  Ropar, Punjab 140001, India \\
  \texttt{2023mcb1308@iitrpr.ac.in} \\
  \And
  Prashant Singh \\
  Department of Mathematics \\
  Indian Institute of Technology Ropar \\
  Ropar, Punjab 140001, India \\
  \texttt{2023mcb1309@iitrpr.ac.in}}
\date{}

\begin{document}
\maketitle

\begin{abstract}
\input{abstract}
\end{abstract}

\input{body}

\clearpage
\bibliography{refs}
\bibliographystyle{acl_natbib}

\clearpage
\input{appendix}

\end{document}

%% file: abstract.tex
Repeated-sampling evaluations increasingly extrapolate pass@\(k\) far beyond
the number \(n\) of samples collected per problem.  We show that, in the
pooled/random-task conditional-Binomial model, fixed-\(n\) success counts
identify only the \(n\) free moments of the latent per-task success
distribution.  Consequently, direct pass@\(k\) is identified for \(k\leq n\),
but generic extrapolated pass@\(k\), tail exponents, and tail constants are
not identified for \(k>n\)---even with arbitrarily many exchangeable tasks at
the same rollout budget.  This is stronger than the observation that the
usual estimator is undefined beyond \(n\): it characterizes the information
missing from the fixed-depth count-law experiment.

We give exact count-law-preserving constructions with incompatible
extrapolations, state the exceptional unique-extension case, and compute
sharp population identified intervals through Hausdorff principal
representations.  On the public 10,000-rollout-per-problem release of Brown
et al., counterfactual \(n=16\) evaluations leave failure at \(k=1000\)
ambiguous by factors from \(1.5\) to over \(2{,}600\) across four
MATH/GSM8K/CodeContests configurations.  The calibration shows that
intermediate-scale failure share alone does not determine width: one GSM8K
configuration is nearly point-identified at \(n=64\), while a controlled
stress test fixes target failure, window share, and hard-task dispersion yet
changes sharp width by orders of magnitude.  The same calibration separates a
raw zero-atom reference from a
Jeffreys-smoothed latent-probability convention: under the latter, the
Beta--Binomial forecast understates residual GSM8K failure by at least
\(85\times\).  Our result does not reject parametric inference-time scaling
laws; it supplies the nonparametric baseline against which their assumptions
can be evaluated.  We give an exact, conservative one-coordinate finite-task
confidence certificate and a reporting standard separating direct estimates,
identified sets, and model-conditioned forecasts.

%% file: body.tex
\section{Introduction}

Repeated sampling can substantially improve LLM coverage on mathematical,
programming, and safety evaluations.  Brown et al.~\citeyearpar{brown2024monkeys}
fit exponentiated power laws to these curves over large sampling budgets.
Schaeffer et al.~\citeyearpar{schaeffer2025power} explain aggregate power-law
failure through a left power-law tail of per-problem pass@1 probabilities.
Most recently, Kazdan et al.~\citeyearpar{kazdan2025efficient} criticize common
curve fits, advocate Beta--Binomial extrapolation, and use adaptive sampling
to forecast pass@\(k\) far beyond the observed per-problem budget.

These are important practical methods, but they invite a prior question.  If
an evaluation observes only \(n\) rollouts per problem, what does its complete
success-count distribution identify about pass@\(k\) for \(k>n\)?  The answer
is not merely that the familiar estimator is undefined there.  Fixed-\(n\)
counts identify only \(n\) free moments of the latent task-difficulty
distribution.  Thus, even with arbitrarily many benchmark problems at the
same rollout budget, generic extrapolated pass@\(k\) remains a partially
identified functional.

This distinction matters directly for the inference-time scaling literature.
Schaeffer et al.'s left-tail premise concerns an unobserved mixing
distribution; a fitted Beta--Binomial model similarly supplies unobserved
moments.  These may be useful modeling assumptions, but they are not
discoveries recovered nonparametrically from fixed-\(n\) counts.  We quantify
the gap rather than arguing against parametric forecasting: a published
forecast should be read alongside the sharp set of all extrapolations
compatible with its fixed-rollout count law.

\paragraph{Contributions.}
\begin{enumerate}
    \item We give a general identification theorem: fixed-\(n\) Binomial
    counts identify expectations of every polynomial functional of the latent
    success probability of degree at most \(n\).  For an interior truncated
    Hausdorff moment sequence, extrapolated pass@\(k\) has a nondegenerate
    identified interval for every \(k>n\); rank-deficient boundary sequences
    can instead uniquely determine their mixing law.  Direct pass@\(k\) for
    \(k\leq n\) is a corollary.
    \item We give two exact constructions with identical count laws and
    incompatible extrapolations.  One separates polynomial from exponential
    failure; the other holds the left-tail exponent fixed while changing its
    leading constant.  For even \(n\), we compute sharp identified intervals
    with Hausdorff principal representations, rather than a support-grid
    approximation.
    \item We calibrate the result on the public 10,000-rollout-per-problem
    release of Brown et al.  In four MATH/GSM8K/CodeContests configurations, the
    sharp \(n=16\) interval at \(k=1000\) ranges from \(1.5\times\) to over
    \(2{,}600\times\) in failure probability.  At \(n=64\), raw plug-in sets
    can be nearly point-like when the full count law isolates the component
    carrying target-budget failure.  Separating the raw zero-frequency
    reference from a Jeffreys-smoothed convention materially changes the
    calibration: conditional Beta--Binomial forecasts are 4.5--29\% above
    the upper posterior endpoint on three configurations and understate
    residual GSM8K failure by more than \(85\times\).  A controlled synthetic
    stress test additionally shows that intermediate-window mass and hard-task
    dispersion, even jointly, cannot characterize sharp-set width: the
    remaining truncated-moment geometry can change it by orders of magnitude.
\end{enumerate}

\section{Setting and directly relevant work}
\label{sec:setting}

Fix a benchmark, prompt, decoding distribution, resource limit, answer
extraction rule, and scorer.  Let \(Y_{ij}\in\{0,1\}\) denote the complete
evaluation-pipeline outcome for completion \(j\) of task \(i\).  We use the
standard mixture model
\[
 Y_{ij}\mid p_i\overset{\mathrm{iid}}{\sim}\operatorname{Bernoulli}(p_i),
 \qquad C_i=\sum_{j=1}^{n}Y_{ij}.
 \tag{1}
\]
Here \(p_i\) is operational: changing a cap, parser, or scorer changes the
quantity being measured.  Conditional iid sampling is the usual premise of
count-based pass@\(k\) evaluation.  It is also implied by an infinitely
exchangeable completion sequence through de Finetti's representation
\citep{hewitt1955symmetric}; finite exchangeability alone is not enough.

Our primary identification experiment is the random-task (or pooled-count)
experiment: \(p_i\stackrel{\mathrm{iid}}{\sim}F\), after which only the
marginal count law is observed.  Equivalently, \(F\) can be a stated
empirical mixing reference used to generate a pooled count law.  Mean failure
and coverage are
\[
 R_F(k)=\mathbb E_F[(1-p)^k],\quad
 \mathrm{pass@}k=1-R_F(k).
 \tag{2}
\]
For \(k\leq n\), the usual direct estimator is
\[
 \widehat R(k)=\frac1N\sum_{i=1}^N
 \frac{\binom{n-C_i}{k}}{\binom nk}.
 \tag{3}
\]
Conditional on the \(p_i\)'s, it is unbiased for \(N^{-1}\sum_i(1-p_i)^k\).
This is the estimator introduced for code-generation pass@\(k\) by
\citet{chen2021codex}; it is a design-based estimate, not an extrapolation
model.

\paragraph{A fixed labeled benchmark is different.}
If one treats \((p_1,\ldots,p_N)\) as an unrestricted but fixed finite
parameter vector and retains its \emph{labeled} product likelihood, that
finite-dimensional likelihood is formally injective: each Binomial marginal
identifies its own \(p_i\).  We do not call
\(N^{-1}\sum_i(1-p_i)^k\) nonidentified in that product model.  Our theorem
instead concerns the information in the pooled/random-task count law, the
asymptotic experiment obtained by increasing the number of exchangeable tasks
while holding the per-task depth fixed.  Under that experiment the counts are
iid from one \(q\), and two mixing laws with the same \(q\) induce exactly the
same data distribution for every number of tasks.  The fixed-benchmark
calibration in Section~\ref{sec:public} uses its empirical distribution only
as a declared reference for this pooled-count experiment.

\paragraph{Where this paper sits.}
Brown et al.~\citeyearpar{brown2024monkeys} fit an exponentiated power law to
repeated-sampling coverage.  Schaeffer et al.~\citeyearpar{schaeffer2025power}
derive aggregate polynomial failure from a left power-law tail in \(F\).
Kazdan et al.~\citeyearpar{kazdan2025efficient} show weaknesses of log--log
and discretized-Beta fits, propose a Beta--Binomial likelihood, and allocate
additional samples to hard problems.  Liu~\citeyearpar{liu2026twocalls}
derives sharp two-moment regions for majority-vote accuracy; our setting uses
arbitrary rollout depth and the at-least-one-success pass@\(k\) functional.
Hariri et al.~\citeyearpar{hariri2026dontpass} instead propose a Bayesian
Dirichlet evaluation and ranking procedure, while Chen et al.~\citeyearpar{chen2026prompts}
study partial identification of a latent label under LLM misclassification.
These works share the premise that repeated-query design governs inference,
but their estimands and observable structures differ from a Binomial mixture
over task-level pass probabilities.

The estimator-domain observation that equation~(3) requires \(k\leq n\)
appears in the scaling literature.  Our added claim is stronger but model
specific: in the pooled/random-task count-law experiment, the pass@\(k\)
functional is generically partially identified beyond \(n\), even in the
infinite-\(N\) limit.  Therefore a Beta--Binomial or tail-law forecast is
conditional on its mixing model, however well it predicts held-out data.

Mathematically, the analysis is a truncated Hausdorff moment problem
\citep{hausdorff1921} and a partial-identification problem
\citep{manski2003partial}.  Principal representations and moment--SOS
methods provide sharp generalized-moment bounds
\citep{krein1977,lassere2001}.  Our contribution is to connect those objects
to the exact target and released data products of multi-rollout LLM
evaluation.

\section{The fixed-rollout identification boundary}
\label{sec:boundary}

Let \(\mathcal P_n\) be the polynomials of degree at most \(n\), and write
\(q_c=\Pr(C=c)=\int \binom nc p^c(1-p)^{n-c}\,dF(p)\).

\begin{theorem}[Fixed-rollout boundary]
\label{thm:boundary}
For fixed \(n\), the count law \(q=(q_0,\ldots,q_n)\) identifies
\(\int h(p)\,dF(p)\) for every \(h\in\mathcal P_n\).  Equivalently, it
identifies \(m_j=\int p^j\,dF(p)\) for \(j=0,\ldots,n\), of which \(n\)
are free because \(m_0=1\).  Hence \(R_F(k)\) is identified for every
integer \(k\leq n\).

For every \(k>n\), there are two distributions on \([0,1]\) with identical
complete count laws and different \(R_F(k)\).  If the induced truncated
Hausdorff moment vector lies in the interior of its moment space, then the
sharp identified interval for \(R_F(k)\) has strictly positive width.
Boundary sequences require separate treatment: some have a unique
representing measure, in which case every higher-order functional is pinned.
\end{theorem}

The Bernstein kernels in \(q\) span \(\mathcal P_n\), while
\[
 R_F(k)=\sum_{j=0}^{k}(-1)^j\binom kjm_j.
 \tag{4}
\]
\paragraph{Proof sketch.}
The \(n+1\) Bernstein kernels in \(q\) form a basis of \(\mathcal P_n\), so
the count law and the first \(n\) free moments contain exactly the same
information.  The shifted-Legendre construction in
Appendix~\ref{app:proofs} gives an explicit observationally equivalent pair
for every \(k>n\).  More generally, the strict-width claim for an arbitrary
interior truncated Hausdorff sequence follows from generalized-moment theory:
\[
 \begin{aligned}
 \frac{d^{n+1}}{dp^{n+1}}(1-p)^k
 &=(-1)^{n+1}\frac{k!}{(k-n-1)!}\\[-2pt]
 &\quad\cdot(1-p)^{k-n-1}.
 \end{aligned}
\]
has constant nonzero sign on \((0,1)\).  Thus
\(1,p,\ldots,p^n,(1-p)^k\) is a strict extended Chebyshev system, giving
distinct generalized-moment extrema in the interior.  Appendix~\ref{app:proofs}
gives the construction, endpoint orientation, and boundary cases.
Operationally, in the pooled/random-task experiment, more tasks make \(q\)
more precise, whereas more rollouts per task enlarge the set of directly
identified functionals; they are not substitutes.

\subsection{Two exact incompatible extrapolations}

\paragraph{Left-tail convention.}
We say that \(F\) has left-tail exponent \(\alpha>0\) and leading constant
\(C>0\) when \(F([0,x])\sim Cx^\alpha\) as \(x\downarrow0\).  If its density
satisfies \(f(p)\sim C\alpha p^{\alpha-1}\), then a standard beta-integral
argument gives \(R_F(k)\sim C\Gamma(\alpha+1)k^{-\alpha}\).  Thus this
left-tail convention is precisely the one underlying asymptotic power-law
failure curves.

\begin{figure*}[t]
\centering
\includegraphics[width=.88\textwidth]{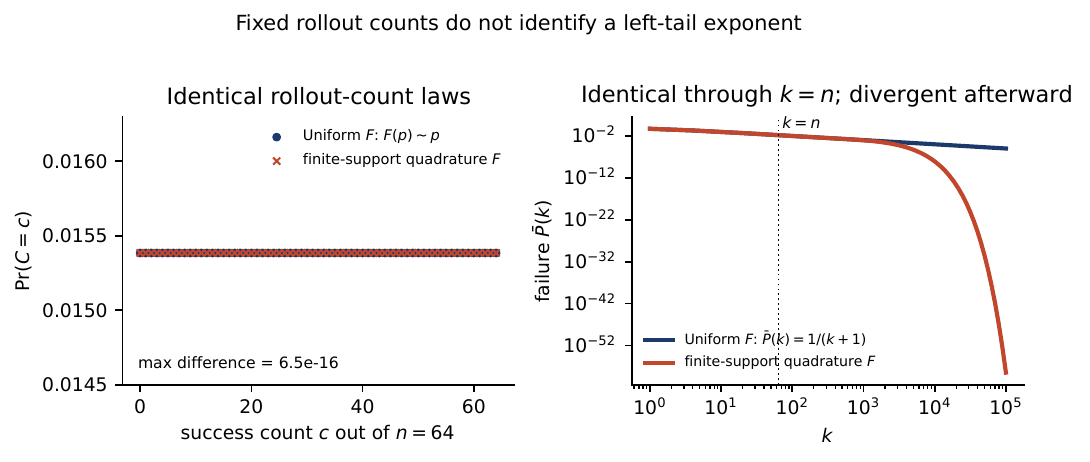}
\caption{Uniform and finite-support mixing laws have identical \(n=64\)
success-count distributions, but incompatible extrapolated failure curves.
No number of tasks at the same fixed rollout budget distinguishes them.}
\label{fig:identifiability}
\end{figure*}

Gauss--Legendre quadrature gives a strong certificate.  Uniform\([0,1]\)
mixing and a positive finite-support quadrature measure integrate every
degree-\(n\) polynomial identically.  They therefore have exactly the same
fixed-\(n\) count law.  But Uniform mixing has polynomial failure
\(1/(k+1)\), while the finite-support law has eventual exponential failure.
Figure~\ref{fig:identifiability} shows the construction at \(n=64\): the
count laws agree to machine precision despite their divergent extrapolations.

A complementary, weaker construction isolates the tail constant.  Let
\(P_{n+1}\) be the Legendre polynomial of degree \(n+1\), and set
\[
 f_\varepsilon(p)=1+\varepsilon P_{n+1}(2p-1),\qquad |\varepsilon|<1.
 \tag{5}
\]
Orthogonality preserves every degree-\(n\) Binomial kernel, while
\[
 F_\varepsilon(x)\sim[1+\varepsilon(-1)^{n+1}]x\quad(x\downarrow0).
 \tag{6}
\]
For even \(n=64\), \(\varepsilon=.8\) gives the same count law and the
same tail exponent \(\alpha=1\), but constants \(C=.2\) and \(C=1\).
Appendix~\ref{app:proofs} gives the closed-form integral, an analytic-density
pair with distinct exponents, and the failure-ratio limit.  Thus the
quadrature construction witnesses qualitatively incompatible tails, whereas
the Legendre family witnesses a bounded multiplicative ambiguity even after
fixing \(\alpha\).

\section{Sharp partial identification and finite-sample inference}
\label{sec:partial}

For a known count law, the population identified set is
\begin{align}
 \mathcal I_{n,k}(q)
 &=\bigl\{R_F(k):F\in\mathcal P([0,1]),\notag\\[-3pt]
 &\quad \Pr_F(C=c)=q_c,\quad c=0,\ldots,n\bigr\}.
 \tag{7}
\end{align}
It is a compact interval.  For even \(n=2r\) and an interior moment
sequence, its endpoints for \((1-p)^k\) are attained by the two Hausdorff
principal representations: the \(r+1\)-atom rules containing, respectively,
\(0\) and \(1\).  We compute those rules by Gaussian quadrature for
\(p\,dF\) and \((1-p)\,dF\).  This is a sharp continuous solution, not a
grid-restricted sensitivity range; the equivalent generalized moment program
also admits a moment--SOS formulation \citep{krein1977,lassere2001}.
Appendix~\ref{app:proofs} gives the endpoint orientation.  For odd rollout
budgets, the same generalized-moment problem uses the lower and upper
Hausdorff principal representations rather than the two single-endpoint
rules.

The qualification about boundary count laws matters.  A flat, rank-deficient
Hausdorff moment sequence can have a unique atomic extension, in which case
\(\mathcal I_{n,k}(q)\) collapses even when \(k>n\).  Our theorem is thus a
generic limitation, not the false claim that every finite count histogram
must imply ambiguity.

\paragraph{Finite samples.}
An empirical histogram need not itself be an exact Binomial-mixture law, so a
plug-in sharp interval is not a confidence interval.  Finite-sample inference
should start with a simultaneous confidence region \(\mathcal Q_N\) for the
count law \(q\), not independent absolute-error bands on high raw moments:
\(m_j=\sum_c (c)_j/(n)_j\,q_c\) is an exact linear map.  The desired
population-set projection is
\(\bigcup_{q\in\mathcal Q_N}\mathcal I_{n,k}(q)\).  One must also state
whether coverage is intended for the scalar \(R_F(k)\) or for the entire
identified set; these are distinct targets in partial-identification
inference \citep{imbensmanski2004,chernozhukovhongtamer2007}.  We also give
one exact but deliberately conservative finite-task certificate.  Let
\(Z=\sum_i\mathbf{1}\{C_i=0\}\), and let \([a,b]\) be a
\(1-\delta\) Clopper--Pearson interval for
\(q_0=\Pr(C=0)=R_F(n)\) \citep{clopperpearson1934}.  For \(k\geq n\),
\[
  a^{k/n}\ \leq\ R_F(k)\ \leq\ b
  \tag{8}
\]
has the same finite-sample coverage: with \(X=(1-p)^n\), Jensen gives
\(R_F(k)=\mathbb E[X^{k/n}]\geq q_0^{k/n}\), while \(X^{k/n}\leq X\).
Appendix~\ref{app:inference} implements this certificate in a prospective
precision calibration.  It uses only \(q_0\), so it is not a replacement for the
sharper complete-count-law projection \(\bigcup_{q\in\mathcal Q_N}
\mathcal I_{n,k}(q)\); rather, it makes the finite-task uncertainty that
such a projection must confront explicit.

\paragraph{Adaptive allocation.}
Dynamic sampling can be beneficial, but it changes the observation model:
outcome-dependent \(n_i\)'s are not summarized by one fixed-\(n\) count law.
Under a maintained Beta--Binomial likelihood they can be modeled, as in
\citet{kazdan2025efficient}.  Under misspecification, allocation and
extrapolation should be assessed jointly with a design-aware sensitivity
analysis; sampling hard tasks more often cannot by itself make a generic
large-\(k\) target nonparametrically identified.

\section{Public 10,000-rollout calibration}
\label{sec:public}

We use the public \emph{Monkey Business} release from
\citet{brown2024monkeys}: per-completion correctness labels for 10,000
samples on each task.  We analyze all 128 released MATH problems for
Llama-3-8B-Instruct and Llama-3-70B-Instruct, all 127 released GSM8K problems,
and all 140 released CodeContests problems for Llama-3-70B-Instruct.  This is
not a new model run.  These are four deliberate case studies, not an
exhaustive or randomly sampled survey of the broader release: they contrast
two MATH model sizes with two qualitatively different non-MATH count-law
geometries.  Accordingly, we make no prevalence claim about width or about a
low-dimensional difficulty diagnostic.  Within each stated case study, every
eligible released task is retained.

For each task, the raw frequency \(\hat p_i=c_i/M_i\) defines a
finite-pool empirical-mixing reference.  For \(n\in\{16,64\}\), we form the
large-\(N\) counterfactual count law
\(N^{-1}\sum_i\operatorname{Bin}(n,\hat p_i)\).  This deliberately removes
ordinary finite-task noise: it asks what an arbitrarily large \(n\)-rollout
pooled-count evaluation could identify under this stated reference.  It is
not a claim that the labeled finite-benchmark product likelihood is
nonidentified.

Raw frequencies and a continuous latent-probability convention are different
estimands, especially when a task has \(c_i=0\).  We therefore do \emph{not}
attach an uncertainty interval to the raw reference.  Separately, we report
the Jeffreys-smoothed reference
\(\tilde p_i=(c_i+1/2)/(M_i+1)\) and the 95\% posterior interval from
independent \(\operatorname{Beta}(c_i+1/2,M_i-c_i+1/2)\) draws, conditional on
the fixed task set.  In CodeContests, 87 of 140 tasks have \(c_i=0\): the
raw reference is .713 whereas the Jeffreys-smoothed reference is .678.
This is a convention sensitivity, not a raw-reference error bar.  We fit
the conditional Beta--Binomial forecast to the same count law under each
convention.

\paragraph{A scale heuristic and its limit.}
A probability \(p\) changes an \(n\)-rollout count law weakly when \(np\)
is small, but affects \(R_F(k)\) materially when \(kp\) is order one.
This exposes the familiar intermediate scale
\(1/k\lesssim p\lesssim1/n\).  Failure mass in that band is therefore a
natural warning signal for extrapolation, but it is not an identification
criterion.  Sharp-set width is determined by the full truncated-moment
geometry.  The number and separation of failure-carrying components are
descriptive features of that geometry, but neither is sufficient by itself.
Under a raw
empirical reference, explicit zero-frequency atoms also contribute directly
to the stated reference failure; this is a convention, not a claim that zero
probability is identified from the count law.

\begin{figure*}[t]
\centering
\includegraphics[width=.94\textwidth]{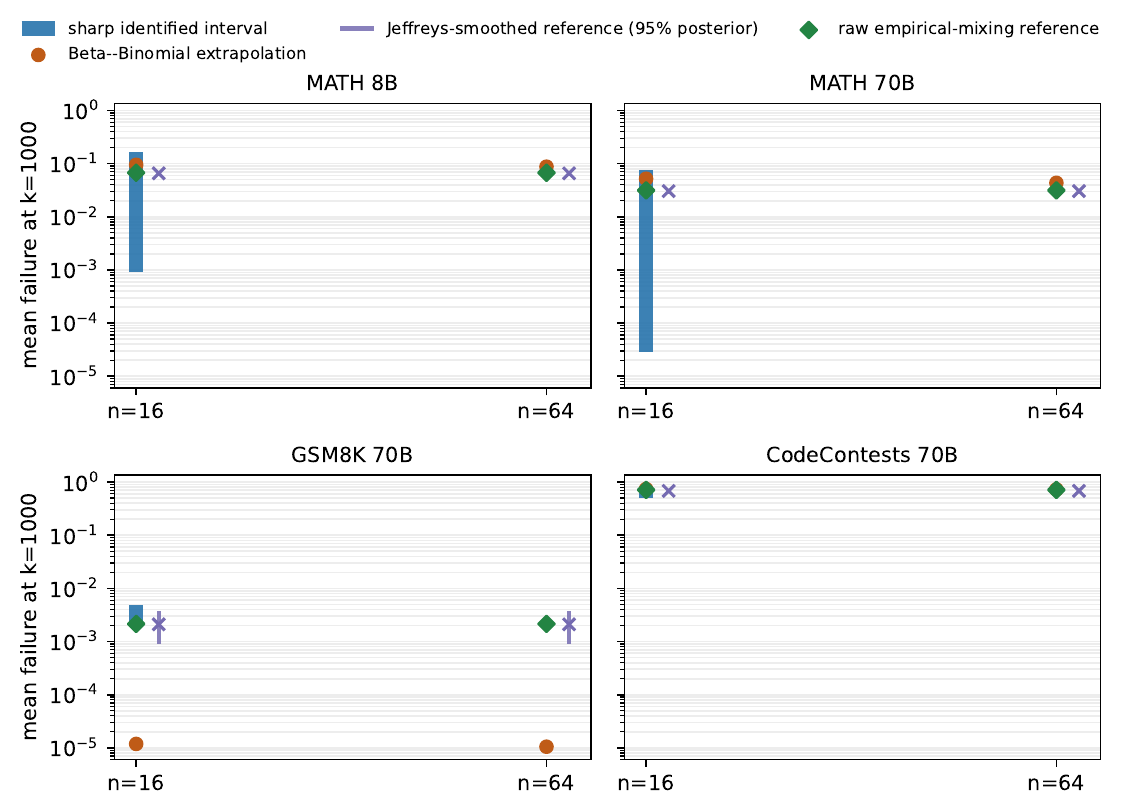}
\caption{Public calibration at \(k=1000\).  Blue bars and orange circles use
the raw \(\hat p_i=c_i/M_i\) convention.  Green diamonds are its finite-pool
empirical-mixing reference.  Purple crosses and bars are the \emph{different}
Jeffreys-smoothed reference and its fixed-task 95\% posterior interval.
Blue intervals are population plug-in sets, not finite-sample confidence
intervals.}
\label{fig:public}
\end{figure*}

\begin{table*}[t]
\centering
\caption{Raw-empirical-reference calibration at \(k=1000\).  \(R_{\rm
raw}(n)\) is directly identified.  Brackets are sharp population plug-in
sets conditional on the raw induced count law.  The last column is the
relative \(n=64\) width \((U-L)/L\), computed at 120-decimal arithmetic;
it describes this stated empirical reference, not a confidence interval or
exact-rank certificate.}
\label{tab:public}
\small
\resizebox{\textwidth}{!}{%
\begin{tabular}{lccccc}
\toprule
Configuration & \(R_{\rm raw}(16)\) & \(R_{\rm raw}(1000)\) & \(n=16\) sharp set & \(n=64\) sharp set & relative \(n=64\) width \\
\midrule
MATH, Llama-3-8B & .331 & .0673 & [\(.000916,.168\)] & [\(.0663,.0683\)] & \(3.00\!\times\!10^{-2}\) \\
MATH, Llama-3-70B & .193 & .0314 & [\(2.89\!\times\!10^{-5},.0755\)] & [\(.0308,.0319\)] & \(3.59\!\times\!10^{-2}\) \\
GSM8K, Llama-3-70B & .00795 & .00214 & [\(.00214,.00476\)] & [\(.00214,.00214\)] & \(8.84\!\times\!10^{-47}\) \\
CodeContests, Llama-3-70B & .842 & .713 & [\(.503,.775\)] & [\(.713,.713\)] & \(2.93\!\times\!10^{-12}\) \\
\bottomrule
\end{tabular}
}
\end{table*}

\begin{table*}[t]
\centering
\caption{Jeffreys-smoothed calibration at \(k=1000\), a distinct
latent-probability convention from Table~\ref{tab:public}.  The interval is
a fixed-task 95\% posterior propagation interval, and the Beta--Binomial
prediction is fit to the same Jeffreys-induced \(n=64\) count law.  It is
therefore not an error bar for the raw reference.}
\label{tab:jeffreys}
\small
\resizebox{\textwidth}{!}{%
\begin{tabular}{lccc}
\toprule
Configuration & Jeffreys posterior median [95\% interval] & Beta--Binomial failure & Beta / posterior median \\
\midrule
MATH, Llama-3-8B & .0656 [\(.0596,.0712\)] & .0873 & \(1.33\times\) \\
MATH, Llama-3-70B & .0304 [\(.0268,.0337\)] & .0434 & \(1.43\times\) \\
GSM8K, Llama-3-70B & .00210 [\(.000898,.00380\)] & \(1.00\!\times\!10^{-5}\) & \(0.00477\times\) \\
CodeContests, Llama-3-70B & .681 [\(.670,.691\)] & .723 & \(1.06\times\) \\
\bottomrule
\end{tabular}
}
\end{table*}

Figure~\ref{fig:public} and Table~\ref{tab:public} make the design boundary
visible: \(R_{\rm raw}(16)\) is directly identified, whereas the set opens
at \(k=1000\).  At \(n=16\), the sharp MATH intervals span factors of
\(183\) (8B) and \(2{,}612\) (70B), even after supplying a 10,000-rollout
reference to remove ordinary count-law estimation noise.  Width is strongly
configuration-dependent.  The raw \(n=64\) GSM8K \emph{relative} width is
\(8.84\times10^{-47}\), while the MATH relative widths remain about 3\%.
Appendix~\ref{app:numerics} verifies every displayed endpoint with the
Golub--Welsch implementation at 120 and 300 decimal digits; the maximum
relative discrepancy is \(5.6\times10^{-54}\).

\begin{figure*}[t]
\centering
\includegraphics[width=.91\textwidth]{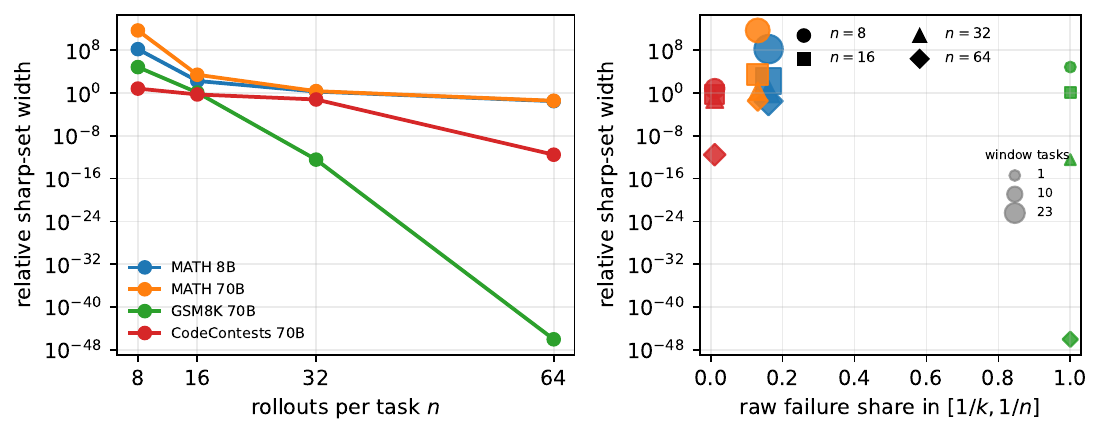}
\caption{Sharp-width trajectories at \(k=1000\) under the raw empirical
reference.  Left: all four calibrations contract as per-task rollout budget
increases.  Right: color denotes configuration, marker shape denotes \(n\),
and marker area denotes the number of window tasks.  GSM8K shows that window
share alone can be maximal while a set is nearly point-like.  These plotted
relationships are descriptive: the complete truncated-moment geometry,
rather than a low-dimensional diagnostic, determines the sharp interval.}
\label{fig:widths}
\end{figure*}

These four calibrations demonstrate that window share alone is insufficient.
At \(n=64\), all of the GSM8K raw-reference failure lies in
\([1/1000,1/64]\), but it is carried by one task with
\(\hat p=13/10{,}000\); all other released GSM8K tasks have
\(\hat p>0.2\).  The full count law separates this isolated component and
the interval contracts already at \(n=32\).  In contrast, the MATH
references distribute target-budget failure over several components and
remain wider.  Figure~\ref{fig:widths} sweeps \(n\in\{8,16,32,64\}\):
the intermediate window is a useful warning scale, but the complete
truncated-moment geometry determines width.

Table~\ref{tab:controlled} makes this qualification constructive.  It holds
fixed a synthetic 128-task hard profile (eight probabilities in
\([.0011,.015]\) plus one at \(.0002\)); only the remaining 119 easy-task
probabilities change.  Every profile has the same target failure to 40
displayed digits, 45.19\% intermediate-window failure share, eight window
tasks, and effective window-task count 2.81.  Nevertheless, the sharp width
varies by a factor of 28 at \(n=16\) and over \(10^{16}\) at \(n=64\).
Thus neither window mass nor a joint window-mass/dispersion summary is a
sufficient characterization of extrapolation ambiguity.

The model comparison must likewise respect the convention.  Under the
Jeffreys convention in Table~\ref{tab:jeffreys}, the Beta--Binomial forecast
is 4.5--29\% above the upper posterior endpoint for MATH and CodeContests,
whereas it is below even GSM8K's lower endpoint by more than \(85\times\).
Thus it can substantially overstate pass@1000.  The apparent agreement with
the CodeContests \emph{raw} reference is partly a zero-frequency-convention
artifact, not a raw-reference uncertainty calculation.  The apparent
worsening from \(n=16\) to \(n=64\) in some rows is not paradoxical: the
nonparametric set can contract faster than model discrepancy vanishes.
These results neither validate nor refute Beta--Binomial forecasting in
general; they quantify its assumption burden in concrete released
evaluations.

\begin{table*}[t]
\centering
\caption{Scale and support diagnostics for the \(n=64\) raw plug-in
calibration.  Window tasks counts empirical probabilities in
\([1/1000,1/64]\); window share is their fraction of raw-reference
failure.  Neither is an identified quantity.  Across these four
illustrative configurations, GSM8K and CodeContests show that either window
share or task count alone can be insufficient.  These are descriptive
summaries, not a theorem or sufficient statistic for sharp-set width.}
\label{tab:sensitivity}
\small
\resizebox{\textwidth}{!}{%
\begin{tabular}{lcccc}
\toprule
Configuration & zero-success tasks & window tasks & window share & relative \(n=64\) width \\
\midrule
MATH, Llama-3-8B & 2 & 23 & 16.1\% & \(3.00\!\times\!10^{-2}\) \\
MATH, Llama-3-70B & 2 & 8 & 13.1\% & \(3.59\!\times\!10^{-2}\) \\
GSM8K, Llama-3-70B & 0 & 1 & 100.0\% & \(8.84\!\times\!10^{-47}\) \\
CodeContests, Llama-3-70B & 87 & 10 & 1.1\% & \(2.93\!\times\!10^{-12}\) \\
\bottomrule
\end{tabular}
}
\end{table*}

\begin{table*}[t]
\centering
\caption{Controlled diagnostic stress test at \(k=1000\).  The hard profile,
target failure (to 40 displayed digits), intermediate-window share (45.19\%),
window-task count (8), and effective window-task count (2.81) are fixed; only
the 119 easy-task probabilities change.  The sharp relative-width range shows
that these coarse summaries do not characterize the full moment geometry.}
\label{tab:controlled}
\small
\begin{tabular}{lcc}
\toprule
Easy-task geometry & relative width at \(n=16\) & relative width at \(n=64\) \\
\midrule
Uniform grid \([.2,.825]\) & \(26.25\) & \(1.09\!\times\!10^{-4}\) \\
Low grid \([.2,.4]\) & \(1.67\) & \(3.97\!\times\!10^{-11}\) \\
Middle grid \([.45,.6]\) & \(0.939\) & \(8.91\!\times\!10^{-21}\) \\
High grid \([.6,.825]\) & \(1.53\) & \(3.21\!\times\!10^{-16}\) \\
Bimodal grids & \(13.48\) & \(2.80\!\times\!10^{-6}\) \\
\bottomrule
\end{tabular}
\end{table*}

\section{What to report and conclude}
\label{sec:protocol}

For direct pass@\(k\), evaluate each task at least \(k\) times and retain the
full evaluation pipeline: because \(p_i\) is operational, changing a cap,
prompt, parser, or scorer changes the target itself.  For \(k>n\), report (i) the exact target and
sampling design, (ii) a population identified-set diagnostic or a
finite-sample confidence region, and (iii) every parametric forecast as a
model-conditioned sensitivity analysis, including its held-out calibration.
Tail exponents and constants require the same separation between fitted and
identified quantities.  In the pooled/random-task experiment, more tasks
improve precision; deeper per-task sampling changes identification.

Simple difficulty summaries can be useful warning signals but should not
replace the identified-set calculation when it is available.  In particular,
the fraction of target-budget failure near \(1/k\lesssim p\lesssim1/n\) does
not determine extrapolation ambiguity; concentration and the remaining moment
geometry can materially alter the sharp bounds.

This also clarifies how to interpret inference-time scaling work.  A power
law or Beta--Binomial forecast can be scientifically useful, especially when
it predicts a held-out rollout budget.  Its evidence is predictive accuracy
under a stated model, not nonparametric recovery of a tail.  The relevant
baseline is therefore not ``no estimate,'' but the compatible interval that
the fixed-rollout design leaves unresolved.

\section{Limitations}

The main theorem is about a pooled/random-task count-law experiment under
conditional iid completions.  Correlated completions, finite exchangeability
without an infinite extension, adaptive task-dependent rollout counts, and
pipeline changes define different observation models.  For a fixed labeled
benchmark, the unrestricted finite product model is formally identified; the
paper's limitation is not that algebraic fact but the information supplied by
a fixed-depth pooled-count design.  The public calibration conditions on
explicit raw-frequency or Jeffreys conventions and is a population plug-in
diagnostic, not a finite-sample confidence interval.  Appendix~\ref{app:inference}
adds an exact but deliberately coarse scalar certificate.  A sharper
finite-sample \emph{identified-set} projection requires a confidence region
for the complete count law and a corresponding projection through the moment
problem; we state that construction but do not claim to have implemented that
stronger procedure here.

\section{Conclusion}

Fixed-rollout pass@\(k\) evaluation is a finite-moment problem.  In the
pooled/random-task model, the direct estimator's \(k\leq n\) domain reflects
a deeper generic identification boundary that more benchmark tasks cannot
cross.  Sharp moment bounds and
public 10,000-rollout calibrations make the remaining assumption burden
visible: at \(n=16\), observed widths range from \(1.5\times\) to over
\(2{,}600\times\), while at \(n=64\) an isolated hard component can be
nearly pinned by one specific full count-law geometry.  The controlled test
shows why no window-mass or dispersion summary substitutes for that geometry.
A raw empirical zero-frequency convention can also materially change a
seemingly well-calibrated forecast.  This provides a constructive standard
for reporting extrapolated LLM evaluation: distinguish direct estimates,
partial-identification bounds, and model-conditioned forecasts rather than
treating them as the same kind of evidence.

%% file: appendix.tex
\appendix
\section{Proofs and implementation details}
\label{app:proofs}

\subsection{Direct estimator and polynomial identification}

For a task with success probability \(p_i\), choosing \(k\) failed items
from the \(n-C_i\) observed failures gives
\[
 \mathbb E\left[\frac{\binom{n-C_i}{k}}{\binom nk}\middle|p_i\right]
 =(1-p_i)^k,\qquad k\leq n.
\]
This proves equation~(3).  The success factorial moments obey
\[
 \begin{aligned}
 \mathbb E\left[\frac{(C)_j}{(n)_j}\right]
 &=\int p^j\,dF(p)\\[-2pt]
 &=m_j,\qquad j=0,\ldots,n.
 \end{aligned}
 \tag{A1}
\]
The degree-\(n\) Bernstein polynomials
\(\binom nc p^c(1-p)^{n-c}\) form a basis of \(\mathcal P_n\).  Hence the
count law identifies every integral against a degree-\(n\) polynomial, and
the first \(n\) moments determine the count law conversely.  Equation~(4)
then proves the pass@\(k\) corollary.

\subsection{Legendre perturbations}

Put \(\widetilde P_m(p)=P_m(2p-1)\).  For every polynomial \(h\) of degree
less than \(m\), shifted-Legendre orthogonality gives
\[
 \int_0^1h(p)\widetilde P_m(p)\,dp=0.
 \tag{A2}
\]
Thus equation~(5) preserves all fixed-\(n\) count probabilities.  For
\(k\geq m\), the closed-form integral is
{\small
\[
 \begin{aligned}
 \int_0^1(1-p)^k\widetilde P_m(p)\!dp
 &={}\\[-2pt]
 &\quad(-1)^m\frac{(k!)^2}{(k-m)!(k+m+1)!}.
 \end{aligned}
 \tag{A3}
\]
}
To see this, substitute \(x=1-p\), use
\(\widetilde P_m(1-x)=(-1)^m\widetilde P_m(x)\), and apply the standard
shifted-Legendre monomial integral.  Taking \(m=n+1\) makes (A3) nonzero for
every \(k>n\), proving Theorem~\ref{thm:boundary}'s counterexample.

For the Uniform base law, \(R_{F_0}(k)=1/(k+1)\).  Dividing the perturbation
in (A3) by this quantity and using the ratio of factorials gives
\[
 \frac{R_{F_\varepsilon}(k)}{R_{F_0}(k)}
 \longrightarrow 1+\varepsilon(-1)^{n+1}.
 \tag{A4}
\]
So this family produces the bounded multiplicative distortion matching its
tail-constant ratio.  It is deliberately complementary to the
Gauss--Legendre construction, which produces a qualitative polynomial versus
exponential discrepancy.

The exponent ambiguity is not an artifact of comparing a continuous law with
a finite-support rule.  Let \(m=n+1\),
\(h_m(p)=(-1)^m\widetilde P_m(p)\), and define
\[
 f_0(p)=2p,\qquad f_\varepsilon(p)=2p+\varepsilon h_m(p).
 \tag{A5}
\]
The same orthogonality preserves all moments through \(n\).  Moreover,
\(h_m(0)=1\), \(|h_m|\leq1\), and
\(|h_m'(p)|\leq m(m+1)\).  Hence for
\(0<\varepsilon\leq1/[m(m+1)]\), the density \(f_\varepsilon\) is
nonnegative on \([0,1]\): it is nonnegative near zero because
\(h_m\geq0\) there, and elsewhere the \(2p\) term dominates a negative
perturbation.  Both densities are analytic, but
\[
 F_0([0,x])=x^2,\qquad F_\varepsilon([0,x])=\varepsilon x+O(x^2).
 \tag{A6}
\]
They therefore have exponents 2 and 1 while retaining exactly the same
fixed-\(n\) count law.  This supplies an analytic-density version of the
exponent nonidentification result.

\subsection{Boundary moment sequences and sharp bounds}

The set of representing probability measures for a truncated Hausdorff moment
sequence is compact and convex.  Its image under \(F\mapsto R_F(k)\) is a
compact interval.  An endpoint has an atomic optimizer; for an interior
sequence with moments through \(2r\), the two principal representations
containing 0 and 1 attain the extrema for \((1-p)^k\), \(k>2r\).  The needed
Chebyshev-system check is direct:
\[
 \begin{aligned}
 \frac{d^{2r+1}}{dp^{2r+1}}(1-p)^k
 &=-\frac{k!}{(k-2r-1)!}\\[-2pt]
 &\quad\cdot(1-p)^{k-2r-1}<0.
 \end{aligned}
\]
Hence the representation containing 0 is the upper endpoint and the
representation containing 1 the lower endpoint \citep{krein1977}.  We obtain
their non-endpoint nodes and weights together by Golub--Welsch eigendecomposition
of the Jacobi matrix for the positive measures \(p\,dF\) and \((1-p)\,dF\).
This avoids a raw-power Vandermonde solve.  Appendix~\ref{app:numerics}
reports a precision sweep and deterministic controls; the 120- and
300-decimal endpoint values agree for every public interval used in the
paper.

This interior qualification cannot be dropped.  If the appropriate Hankel
moment matrix is rank deficient and admits a flat extension, the representing
measure can be unique.  Then every higher moment is identified and the
interval is a point.  We do not infer a flat extension from numerical
conditioning, nor do we infer exact zero success probability from an observed
zero count.  The raw zero-frequency convention is stated explicitly in
Section~\ref{sec:public}; the theorem's generic qualification is essential.

\section{Numerical verification}
\label{app:numerics}

The very narrow GSM8K result merits an explicit numerical check.  We rerun
every raw-reference principal-representation calculation at 16, 50, 120, and
300 decimal digits.  At low precision the Cholesky factorization can correctly
report that it cannot resolve the positive moment matrix; we record this
failure rather than force an answer.  Table~\ref{tab:precision} summarizes
the resolution threshold across all four public calibrations.  At 120 and
300 digits all displayed public endpoints agree, with a maximum relative
difference of \(5.6\times10^{-54}\).  This is a precision-stability check,
not an inference claim about the source tasks.

\begin{table}[t]
\centering
\caption{Precision sweep across the four public raw-reference calibrations.
An entry gives the number of configurations for which the interior
factorization resolved; lower precisions that do not resolve return no
endpoint.  Every 120-digit endpoint agrees with its 300-digit counterpart.}
\label{tab:precision}
\small
\begin{tabular}{lrrrr}
\toprule
\(n\) & 16 digits & 50 digits & 120 digits & 300 digits \\
\midrule
8 & 4/4 & 4/4 & 4/4 & 4/4 \\
16 & 4/4 & 4/4 & 4/4 & 4/4 \\
32 & 0/4 & 4/4 & 4/4 & 4/4 \\
64 & 0/4 & 2/4 & 4/4 & 4/4 \\
\bottomrule
\end{tabular}
\end{table}

\begin{table*}[t]
\centering
\caption{Deterministic numerical controls at \(k=1000,n=64\).  The
continuous Uniform control has known reference failure \(1/1001\) and a
genuinely nondegenerate sharp interval.  The synthetic control has 127
distinct atoms, one at \(p=.0013\) and 126 in \([.2,.825]\).  Both are
computed by the same Golub--Welsch code as the public calibration.}
\label{tab:numerics}
\small
\begin{tabular}{lccc}
\toprule
control & reference failure & sharp interval & relative width \\
\midrule
Uniform\([0,1]\) & .000999 & [\(.000909,.00111\)] & .223 \\
127-atom single-hard-task profile & .00214 & [\(.00214,.00214\)] & \(1.01\!\times\!10^{-22}\) \\
\bottomrule
\end{tabular}
\end{table*}

The Uniform calculation rules out a generic collapse of the implementation:
the same \(n=64\) code returns a 22.3\% interval.  Conversely, the synthetic
control establishes only feasibility: an extremely narrow interval can occur
for a nondegenerate empirical law when one separated hard component carries
the target-budget failure.  It is not designed to reproduce GSM8K's
\(8.84\times10^{-47}\) width; that number depends on the detailed locations
of its 126 easy-task frequencies.  In the released GSM8K calibration the
only empirical probability below .2 is \(13/10{,}000\).

\section{Exact finite-sample certificate}
\label{app:inference}

For an actual \(N\)-task random-task evaluation, \(Z=\sum_i
\mathbf{1}\{C_i=0\}\sim\operatorname{Binomial}(N,q_0)\), where
\(q_0=R_F(n)\).  Let \([a(Z),b(Z)]\) be the two-sided
\(1-\delta\) Clopper--Pearson interval for \(q_0\).  The event
\(q_0\in[a(Z),b(Z)]\) has probability at least \(1-\delta\).  Setting
\(X=(1-p)^n\in[0,1]\) and \(s=k/n\geq1\), Jensen and pointwise
monotonicity imply
\[
  \begin{aligned}
  R_F(k)&=\mathbb E[X^s]\geq\mathbb E[X]^s=q_0^s,\\
  R_F(k)&=\mathbb E[X^s]\leq\mathbb E[X]=q_0.
  \end{aligned}
\]
Thus \([a(Z)^{k/n},b(Z)]\) is an exact, distribution-free finite-sample
confidence interval for the scalar \(R_F(k)\).  It is intentionally coarse:
it uses only \(q_0\), rather than the full histogram.  Its role is to give a
valid executable baseline, not to stand in for the sharp
\(\mathcal Q_N\)-projection described in Section~\ref{sec:partial}.

\begin{table*}[t]
\centering
\caption{Prospective finite-task precision calibration of the exact 95\%
Clopper--Pearson/Jensen certificate at \(n=16,k=1000\).  Each of 20,000
simulated \(N\)-task count evaluations draws \(p_i\) with replacement from
the stated raw empirical-mixing reference and then \(C_i\sim
\operatorname{Binomial}(16,p_i)\).  The displayed certificate is its componentwise median across
simulations.  All 20,000 simulated targets are covered; this empirical
frequency is a check, whereas the certificate's at-least-95\% coverage is
the analytic guarantee.  These are prospective precision calculations under the
raw convention, not uncertainty intervals for the 10,000-rollout frequencies.}
\label{tab:finite}
\small
\resizebox{\textwidth}{!}{%
\begin{tabular}{lcc}
\toprule
Configuration & raw-reference \(R_F(1000)\) & median exact 95\% certificate \\
\midrule
MATH, Llama-3-8B & .0673 & [\(1.34\!\times\!10^{-38},.417\)] \\
MATH, Llama-3-70B & .0314 & [\(5.48\!\times\!10^{-56},.275\)] \\
GSM8K, Llama-3-70B & .00214 & [\(5.29\!\times\!10^{-232},.0431\)] \\
CodeContests, Llama-3-70B & .713 & [\(9.32\!\times\!10^{-8},.899\)] \\
\bottomrule
\end{tabular}
}
\end{table*}

The exact certificate is wide for all four 100--150-task references.  This
is expected: it is a one-coordinate baseline, while the sharp population
sets condition on a known count law.  The contrast makes the two sources of
uncertainty visible rather than conflating them.  A tighter finite-sample
identified-set interval requires a simultaneous multinomial region
\(\mathcal Q_N\), the exact linear map
\(m_j=\sum_c(c)_j/(n)_j\,q_c\), and the projection
\(\bigcup_{q\in\mathcal Q_N}\mathcal I_{n,k}(q)\).  That full-moment
projection remains an important extension; the certificate above is included
because it gives a correct finite-sample baseline now.

\section{Public-data protocol}

{\raggedright\sloppy
We stream only the Boolean \texttt{is\_corrects} vectors from commit
\texttt{a9f8f73bcd6948a57ed922\allowbreak cba4e48062ef95f553} of the public
\textit{Monkey Business} release.  ``Eligible'' means a source record with a
Boolean \texttt{is\_corrects} vector of length at least 10,000; the
calibration uses every such record: 128 MATH tasks for each Llama-3 model,
127 GSM8K tasks, and 140 CodeContests tasks.  It constructs
\(N^{-1}\sum_i\operatorname{Bin}(n,\hat p_i)\) under each stated frequency
convention and computes the two even-order principal representations at
120-decimal arithmetic.  The independent 10,000-rollout direct estimator is
retained only as a check; the plotted reference is the empirical mixing
functional conditioning the plug-in set.\par}

\paragraph{Beta--Binomial fit.}
For a stated convention and rollout depth \(n\), let
\(q^{(n)}=(q^{(n)}_0,\ldots,q^{(n)}_n)\) denote the induced count law.  Under
\(P\sim\operatorname{Beta}(a,b)\), its model count probabilities are
\[
 \begin{aligned}
 q_c^{\rm BB}(a,b)
 &= {n\choose c}\frac{B(c+a,n-c+b)}{B(a,b)},\\[-2pt]
 &\hspace{4cm} c=0,\ldots,n.
 \end{aligned}
 \tag{A7}
\]
We minimize the count-law cross-entropy
\(-\sum_c q^{(n)}_c\log q_c^{\rm BB}(a,b)\) over \(a,b>0\), using
\(\log a,\log b\) coordinates and log-gamma evaluation.  The implementation
uses a deterministic two-dimensional Nelder--Mead search for 450 iterations
from five starts \((-2,-2),(0,0),(1.5,1.5),(-4,0),(0,-4)\), retaining the
lowest objective.  It then reports the model-conditioned forecast
\[
 R_{\rm BB}(k)=\frac{B(a,b+k)}{B(a,b)}.
 \tag{A8}
\]
This is a reproducible model projection of the induced count law, not an
estimate of a nonparametrically identified tail.

\paragraph{Optimization audit.}
For the 16 raw and Jeffreys fits reported at \(n\in\{16,64\}\), all five
documented starts agree in normalized count-law cross-entropy to within
\(4.85\times10^{-14}\), with maximum absolute forecast variation
\(1.18\times10^{-7}\).  Extending each start by 900 additional
Nelder--Mead iterations changes any forecast by at most
\(2.94\times10^{-8}\).  An independent analytic-gradient BFGS check,
initialized at each retained Nelder--Mead solution, satisfies a
\(10^{-7}\) gradient-norm tolerance and neither lowers the recorded objective
nor changes the forecast at displayed double precision.  The audit records
all starts, objectives, forecasts, and optimizer diagnostics for release.

\paragraph{Reproducibility and release.}
Upon publication we will release the anonymous source task identifiers,
per-task success counts, conventions, moment vectors, solver residuals,
precision checks, Beta--Binomial fit starts and objective values, random
seeds, and the scripts used for every table and figure.  The review manuscript
contains the algorithmic details needed to audit the calculations without a
review-time external link.\par